\documentclass[9pt]{article}
\usepackage{INTERSPEECH2022,amsmath,graphicx}
\usepackage[utf8]{inputenc}
\usepackage{amsmath,graphicx}
\usepackage{multirow}
\usepackage{textcomp}
\usepackage{hyperref}
\hypersetup{colorlinks=true}
\usepackage{moredefs}
\usepackage[final]{changes}

\usepackage{bibspacing}
\usepackage{color}
\definecolor{darkspringgreen}{rgb}{0.39, 0.65, 0.27}

\title{Mixture-of-Expert Conformer for Streaming Multilingual ASR}
\name{Ke Hu, Bo Li, Tara N. Sainath, Yu Zhang, Francoise Beaufays}
\address{Google LLC, USA}
\email{huk@google.com}

\begin{document}
\ninept

\maketitle

\begin{abstract}
End-to-end models with large capacity have significantly improved multilingual automatic speech recognition, but their computation cost poses challenges for on-device applications. We propose a streaming truly multilingual Conformer incorporating mixture-of-expert (MoE) layers that learn to only activate a subset of parameters in training and inference. The MoE layer consists of a softmax gate which chooses the best two experts among many in forward propagation. The proposed MoE layer offers efficient inference by activating a fixed number of parameters as the number of experts increases. We evaluate the proposed model on a set of 12 languages, and achieve an average 11.9\% relative improvement in WER over the baseline. Compared to an adapter model using ground truth information, our MoE model achieves similar WER and activates similar number of parameters but without any language information. We further show around 3\% relative WER improvement by multilingual shallow fusion.
\end{abstract}

\section{Introduction}

An end-to-end (E2E) multilingual automatic speech recognition (ASR) model is appealing because of its potential to recognize multiple languages using a single model. There has been significant effort in multilingual modeling using E2E models \cite{watanabe2017language,kim2018towards,zhou2018multilingual,kannan2019large,hou2020large,zhu2020multilingual,zhou2022configurable}. Previous studies on multilingual ASR have investigated different model structures such as connectionist temporal classification (CTC) models \cite{kim2018towards}, long-short term memory \cite{kannan2019large}, and attention based models \cite{watanabe2017language, zhou2018multilingual,hou2020large, zhu2020multilingual}. Among them, streaming models \cite{zhu2020multilingual, li2022language} are promising candidates for on-device applications. By increasing the model capacity, \cite{li2022language} proposes an on-device streaming multilingual RNN-T model and achieves comparable recognition quality and latency compared to monolingual models. To further increase model capacity, it is crucial to keep computation low for on-device applications. 

Other studies have also shown that increasing model capacity is a key factor in improving performance. By increasing a multilingual E2E model up to 1B size, \cite{pratap2020massively} has improved quality of all variants of the multilingual model. In \cite{li2022massively}, the authors show that under a life-long learning strategy, the model performs consistently better as the capacity increases up to 1B parameters. The same trend has been observed in a two-pass multilingual deliberation model \cite{hu2023scaling}. More recently, the Whisper model \cite{radford2022robust} and Google Universal Speech Model (USM) \cite{zhang2023google} have achieved human-approaching performance with the help of large-scale data and model sizes in billions of parameters.

Larger models come with more cost in training and inference. To improve the modeling efficiency, there have been several approaches in leveraging language-specific components for inference \cite{zhu2020multilingual, mavandadi2023truly, zhang2022streaming, gaur2021mixture}. However, how to predict language information reliably in a streaming fashion is a challenge itself. Others improve efficiency by neural network pruning \cite{yang2022learning} or mixture-of-expert type of models \cite{you2021speechmoe, lu2020bi, wang2020deep, fedus2021switch, you2022speechmoe2} (more discussion in Sect. \ref{sec:related}).

In this work, we propose to use mixture-of-expert (MoE) layers \cite{du2022glam, shazeer2017outrageously} to replace the feed-forward network (FFN) in the Conformer \cite{gulati2020conformer} for multilingual ASR. The MoE layer consists of multiple FFNs and a gating network \cite{shazeer2017outrageously}. The gating network is a softmax over the number of experts, and the outputs of the top two experts are combined in a weighted fashion as the final output. The proposed model is thus sparse when the number of total experts is greater than two. Such an MoE layer has been used in NLP \cite{du2022glam} as well as shallow fusion \cite{hu2023massively} and achieved superior quality compared to their dense counterparts. By adding the MoE layers to the end FFN network of the Conformer layers, we improve the average WER of 12 languages by 11.9\% relative compared to the baseline. In another comparison with a larger baseline (dense model) with a similar total size as the MoE model, we show that the proposed model achieves similar quality by activating only 53\% of parameters during inference. 

We also compare to an adapter model based on \cite{kannan2019large, li2023efficient}. By increasing the total number of experts while activating the top two experts during inference, we achieve similar quality compared to a ground truth language information based adapter model. The two models activate the same number of parameters for inference, however, we note that our model does not need any language information in inference and more straightforward in deployment. Finally, we further improve the MoE model performance by around 3\% relative by shallow fusion using a multilingual neural LM.

\section{Related Work}
\label{sec:related}

There have been several related mixture-of-expert approaches for ASR modeling, but our work differs from them in the following ways. Our MoE study is for streaming multilingual ASR compared to \cite{you2021speechmoe, you2022speechmoe2}, which are for monolingual ASR. We show that our Conformer based MoE structure is effective for multiple languages in Sect. \ref{sec:compare}. Compared to \cite{you2021speechmoe, you2022speechmoe2}, our MoE model also works without using any shared embedding network for expert routing. Other MoE models have been proposed for image processing or natural language processing (NLP) \cite{wang2020deep, fedus2021switch}. In terms of model structure, DeepMoE \cite{wang2020deep} uses embedding network for expert routing and the Switch Transformer \cite{fedus2021switch} activates only one expert layer layer. Further, the performance of DeepMoE and Switch Transformer in ASR is unclear. In multilingual ASR, a mixture of informed-expert model is proposed in \cite{gaur2021mixture}. However, it needs language information to select the expert, and the number of experts increases linearly with the number of languages. In a similar line, a per-language second pass model \cite{mavandadi2023truly} and the adapter model \cite{kannan2019large, li2023efficient} can also be considered as informed-expert models since the language information is used to choose the corresponding module (cascaded encoder or adapter). Compared to \cite{mavandadi2023truly}, the proposed MoE model does not rely on any external language information for routing and is thus more generic. Although \cite{zhang2022streaming} predicts the language information for the second-pass, this increases both training and inference complexity and potential error propagation in practical applications.

In summary, the novelty of our work are mainly two folds: 1) We have proposed an MoE-based Conformer for multilingual ASR and have shown its effectiveness compared to dense models and adapter models, and 2) Our proposed MoE model is relatively simplistic, and does not require language information compared to  \cite{mavandadi2023truly, gaur2021mixture} and any shared embedding network for routing \cite{you2021speechmoe, you2022speechmoe2}.

\section{Conformer with Mixture-of-Experts}
\label{sec:moe}

\begin{figure}[t]
  \centering
  \includegraphics[scale=0.35]{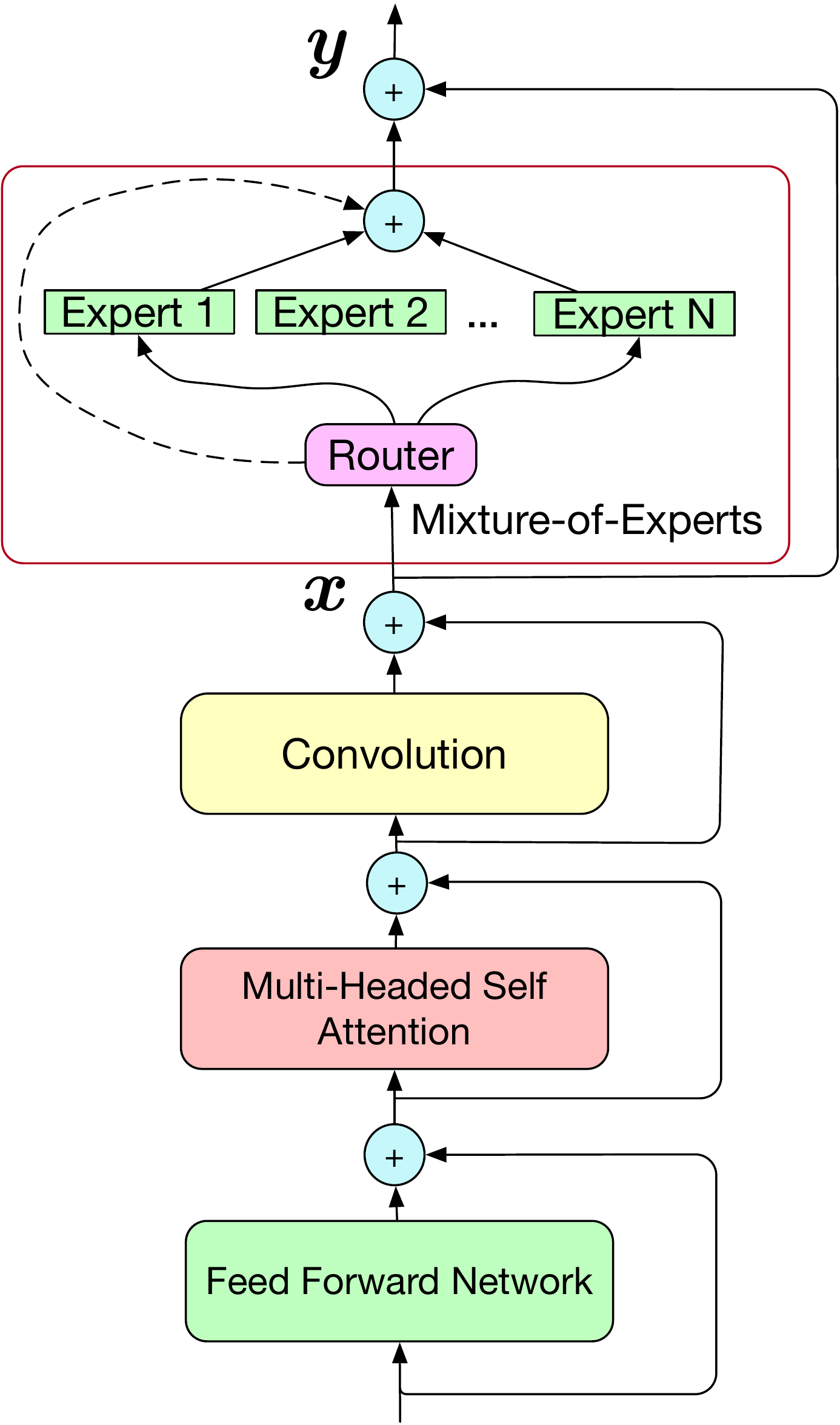}
  \caption{Conformer layer with mixture of experts at the position of the end FFN layer.}
  \label{fig:moe}
  \vspace{-0.3in}
\end{figure}

Our model uses Conformer as the main building block for causal and non-causal encoders. A Conformer layer \cite{gulati2020conformer} consists of a multi-headed self-attention and a convolution-based layer sandwiched by two feed-forward networks (FFN). As shown in Fig. \ref{fig:moe}, to incorporate experts, we use an MoE layer \cite{shazeer2017outrageously, du2022glam} to replace the end FFN in the Conformer layers. Similar to \cite{shazeer2017outrageously, du2022glam}, the MoE layer consists of a routing network and multiple experts, each of which is an FFN. To route a speech frame, we first use a softmax to estimate expert weights: 
\begin{equation}
\vec{g}_{l} = Softmax(W_{l} \cdot \vec{x})
\label{eq:gating}
\end{equation}
where $\vec{x}$ is the output of the previous layer, and $W_l$ is the weight matrix for the router at $l$th Conformer layer. Then the input is routed to the two experts with the highest weights (i.e., top 2 experts) and their outputs are weighted and summed to produce the final output:
\begin{equation}
\vec{y} = \sum_{i=1}^2g_{l,i}*\vec{e}_{l,i}
\label{eq:moe}
\end{equation}
where $g_{l,i}$ is the weight for the top $i$th expert at the $l$th layer. $\vec{e}_{l,i}$ is the corresponding output of the expert. We use the top 2 experts for both training and inference. In Fig. \ref{fig:moe}, we replace the end FFN of the Conformer layer with the MoE layer. We have also tried replacing the start FFN layer or both of them (see more results in Sect. \ref{sec:ablation}).

We use the RNN-T loss \cite{graves2012sequence} for training the Conformer model with MoE layers. To ensure load balance across different experts, we use the same auxiliary loss as in \cite{lepikhin2020gshard}: $ l_{aux} = \frac{1}{N} \sum_{i=1}^N\frac{c_i}{S}\cdot m_i$, where $m_i$ is the average number of times $i$th expert is selected over all frames, and $c_i$ is the expert decision count for $i$th expert derived from the top-2 operation.
We use the mean gates per expert $m_i\cdot(c_i/S)$ as a differentiable approximation for $(c_i/S)^2$ (more details in \cite{lepikhin2020gshard}).

\section{Experiment Setup}

\subsection{Data}
\label{sec:data}

 Our 12-language group consists of English (USA), Mandarin, French, German, Japanese, Spanish (USA), Spanish (Spain), Arabic, Italian, Hindi, Portuguese, and Russian (see Table \ref{tab:data} for more details). Our supervised training data of 12 language locales come from multiple domains such as 
 Voice Search and
 YouTube. In total, they constitute around 139.4M utterances. The training data is anonymized and human transcribed. The per-language number of utterances ranges from 500K to 25.2M. For each language, we use a test set with utterances
 sampled from the Voice Search traffic
ranging from 1.4K to 10K. The test sets do not overlap with the training set, and are also anonymized and human transcribed. We use word error rate (WER) for evaluation, and for languages such as zh-TW, the WER is computed based on characters. We are aware of the sensitive nature of the ASR research and other AI technologies used in this work.
We thus ensure that this work abides by the Google AI Principles \cite{googleai}.
Note that we have removed further details to follow the double blind review process, and will provide them in the camera-ready version.
 
 \begin{table}[t]
\centering
\vspace{-0.1in}
\begin{tabular}{llr}
\toprule
Locale & Language & Counts (M)\\
\midrule
\midrule
en-US & English (USA) & 18.1\\
zh-TW & Mandarin & 0.5\\
fr-FR & French & 10.8\\
de-DE & German &  3.8\\
ja-JP & Japanese & 10.9\\
es-US & Spanish (USA) & 25.2\\
es-ES & Spanish (Spain) & 20.3\\
ar-EG & Arabic & 3.8\\
it-IT & Italian & 13.0\\
hi-IN & Hindi & 14.2\\
pt-BR & Portuguese & 13.4\\
ru-RU & Russian & 5.3\\
\midrule 
\midrule
Total & & 139.4 \\
\end{tabular}
\caption{Training data for 12 language locales. Utterance counts are in millions (M).}
\label{tab:data}
\vspace{-0.3in}
\end{table}

\subsection{Modeling Details}

\subsubsection{Baseline Multilingual Model}
We use a language agnostic multilingual model similar to \cite{li2022language} as the baseline. The baseline model consists of a 7-layer causal Conformer encoder and a 10-layer non-causal cascaded encoder. The causal encoder includes two blocks separated by a stacking layer. The first block consists of an input projection layer and 3 convolution layers. The stacking layer concatenates two neighboring encodings in time to form a 60-ms frame rate. The second block starts with a 1024-dim Conformer layer, and then a projection layer to reduce the model dimension back to 512 for the rest of the causal layers. Note that the causal Conformer layers uses causal convolution and left-context attention
and is thus strictly causal. Secondly, the non-causal layers are cascaded \cite{narayanan2021cascaded} to the causal encoder output. The 10 layers of non-causal Conformer layers have a dimension of 640, and a total right-context of 0.9 sec. We use separate decoders for causal and non-causal encoders to achieve the best quality.

Each transducer decoder consists of a prediction network and a joint network \cite{graves2012sequence}. For the prediction network, we use two embedding layers to embed current and previous tokens separately and concatenate the embeddings as output. The joint network is a single feed-forward layer of 640 units. We use a hybrid autoregressive transducer (HAT) version of the decoder \cite{variani2020hybrid}. A softmax is used to predict 16,384 wordpieces. We generate the wordpieces using mixed transcripts pooled from all languages. The baseline multilingual transducer model has a total of 180M parameters.

\begin{table}[t]
\centering
\begin{tabular}{lcccc}
    \toprule
    & \multicolumn{1}{c}{B1} & \multicolumn{1}{c}{E1} & \multicolumn{1}{c}{E2} & \multicolumn{1}{c}{E3} \\ \cline{2-5}
    \multirow{2}{*}{Model} & \multirow{2.2}{*}{No MoE} & \multirow{2.2}{*}{MoE-Start} & \multirow{2.2}{*}{MoE-End} & \multirow{2.2}{*}{MoE-Both} \\
    & & & \\ \cline{1-5}
    \multirow{1.2}{*}{Total Size} & \multirow{1.2}{*}{180M} & \multirow{1.2}{*}{400M} & \multirow{1.2}{*}{400M} & \multirow{1.2}{*}{640M} \\ [0.3ex] \hline
    \multirow{1.2}{*}{Inf. Size} & \multirow{1.2}{*}{180M} & \multirow{1.2}{*}{211M} & \multirow{1.2}{*}{211M} & \multirow{1.2}{*}{246M} \\ [0.3ex] \hline
    en-US & 9.4 & 8.4 & 8.3 & 8.0 \\
    fr-FR & 10.8 & 9.3 & 9.3 & 8.6 \\
    es-US & 6.8 & 6.0 & 6.0 & 5.6 \\
    es-ES & 6.6 & 5.2 & 4.9 & 4.8 \\
    ja-JP & 15.5 & 13.1 & 12.9 & 12.0 \\
    zh-TW & 9.4 & 9.2 & 9.1 & 9.2 \\
    de-DE & 14.5 & 12.8 & 12.6 & 11.9 \\
    it-IT & 10.4 & 8.8 & 8.8 & 8.1 \\
    ar-EG & 12.3 & 10.9 & 10.8 & 10.3 \\
    pt-BR & 7.6 & 7.0 & 6.9 & 6.5 \\
    ru-RU & 13.0 & 11.0 & 10.9 & 10.4 \\
    hi-IN & 19.6 & 19.5 & 19.2 & 19.1 \\ \hline
    \multirow{1.2}{*}{Avg. WER} & \multirow{1.2}{*}{11.33} & \multirow{1.2}{*}{10.10} & \multirow{1.2}{*}{9.98} & \multirow{1.2}{*}{\textbf{9.54}} \\[0.3ex] \hline
\end{tabular}
\caption{WERs (\%) by placing the MoE layers at different places of the cascaded Conformer encoder.}
\vspace{-0.2in}
\label{tab:wer_nc}
\end{table}

\begin{table}[t]
\centering
\begin{tabular}{lccc}
    \toprule
    & \multicolumn{1}{c}{E2} & \multicolumn{1}{c}{E4} & \multicolumn{1}{c}{E5} \\ \cline{2-4}
    \multirow{2}{*}{Model} & \multirow{2.2}{*}{8-Exp.} & \multirow{2.2}{*}{4-Exp.} & \multirow{2.2}{*}{2-Exp.} \\
    & & & \\ \cline{1-4}
    \multirow{1.2}{*}{Total Size} & \multirow{1.2}{*}{400M} & \multirow{1.2}{*}{295M} & \multirow{1.2}{*}{211M} \\ [0.3ex] \hline
    \multirow{1.2}{*}{Inf. Size} & \multirow{1.2}{*}{211M} & \multirow{1.2}{*}{211M} & \multirow{1.2}{*}{211M} \\ [0.3ex] \hline
    en-US & 8.3 & 8.5 & 8.6 \\
    fr-FR & 9.3 & 9.9 & 9.9 \\
    es-US & 6.0 & 6.2 & 6.3 \\
    es-ES & 4.9 & 5.7 & 5.8 \\
    ja-JP & 12.9 & 14.2 & 14.5 \\
    zh-TW & 9.1 & 8.7 & 9.4 \\
    de-DE & 12.6 & 13.6 & 13.7 \\
    it-IT & 8.8 & 9.2 & 9.2 \\
    ar-EG & 10.8 & 11.2 & 11.3 \\
    pt-BR & 6.9 & 6.8 & 7.2 \\
    ru-RU & 10.9 & 11.3 & 11.5 \\
    hi-IN & 19.2 & 19.5 & 19.5 \\ \hline
    \multirow{1.2}{*}{Avg. WER} & \multirow{1.2}{*}{\textbf{9.98}} & \multirow{1.2}{*}{10.40} & \multirow{1.2}{*}{10.58} \\[0.3ex] \hline
\end{tabular}
\caption{WERs (\%) by reducing the number of MoE layers.}
\label{tab:wer_expert}
\vspace{-0.2in}
\end{table}

\begin{table}[t]
\centering
\begin{tabular}{lccc}
    \toprule
    & \multicolumn{1}{c}{E2} & \multicolumn{1}{c}{E6} & \multicolumn{1}{c}{E7} \\ \cline{2-4}
    \multirow{2}{*}{Model} & \multirow{2.2}{*}{MoE-End} & \multirow{2.2}{*}{MoE-End-Odd} & \multirow{2.2}{*}{MoE-Conf1} \\
    & & & \\ \cline{1-4}
    \multirow{1.2}{*}{Total Size} & \multirow{1.2}{*}{400M} & \multirow{1.2}{*}{295M} & \multirow{1.2}{*}{203M} \\ [0.3ex] \hline
    \multirow{1.2}{*}{Inf. Size} & \multirow{1.2}{*}{211M} & \multirow{1.2}{*}{196M} & \multirow{1.2}{*}{183M} \\ [0.3ex] \hline
    en-US & 8.3 & 8.5 & 8.8 \\
    fr-FR & 9.3 & 10.1 & 10.6 \\
    es-US & 6.0 & 6.4 & 6.5 \\
    es-ES & 4.9 & 5.9 & 6.2 \\
    ja-JP & 12.9 & 14.6 & 14.8 \\
    zh-TW & 9.1 & 8.6 & 8.9 \\
    de-DE & 12.6 & 13.7 & 14.1 \\
    it-IT & 8.8 & 9.3 & 9.9 \\
    ar-EG & 10.8 & 11.4 & 12.0 \\
    pt-BR & 6.9 & 6.8 & 7.2 \\
    ru-RU & 10.9 & 11.3 & 12.0 \\
    hi-IN & 19.2 & 19.4 & 19.5 \\ \hline
    \multirow{1.2}{*}{Avg. WER} & \multirow{1.2}{*}{\textbf{9.98}} & \multirow{1.2}{*}{10.50} & \multirow{1.2}{*}{10.88} \\[0.3ex] \hline
\end{tabular}
\caption{WERs (\%) by reducing the number of MoE layers.}
\label{tab:wer_layer}
\vspace{-0.4in}
\end{table}

\subsubsection{MoE Conformer}
We replace the start, the end, or both FFNs of the Conformer layers by MoE layers (see experiments in Sect. \ref{sec:ablation}) in the cascaded encoder of the baseline model. We use up to 24 experts in our experiments and dynamically choose the top 2 for training and inference. The expert FFNs have the same structure as the Conformer FFNs.  We use the auxiliary loss in Sect. \ref{sec:moe} to encourage load balance between experts. In training, we compute an over-capacity ratio which tracks whether certain experts are overloaded. It is calculated as, for each batch, the percentage of tokens going through a specific expert above a threshold. Our training shows that most experts have over capacity ratios ranging from 0.01 to 0.2 and only 10\% of them range from 0.2 to a maximum value of 0.35, which is quite balanced. The model is trained to predict the same 16,384 wordpieces as the baseline.

We divide the input speech using 32-ms windows with a frame rate of 10 ms. 128D log-Mel filterbank features are extracted from each frame and then stacked together from 3 previous continuous frames to form a 512D input vector. These input vectors are further downsampled to have a 30-ms frame rate. We use SpecAug \cite{park2019specaugment} to improve model robustness against noise. Two frequency masks with a maximum length of 27 and two time masks with a maximum length of 50 are used.

\begin{table*}[t]
\resizebox{\textwidth}{!}{%
\begin{tabular}{|ccccccccccccccccc|}
    \hline
    \multirow{2}{*}{ID} & \multirow{2}{*}{Model} & \multirow{2}{*}{\shortstack{Total \\Size}} & \multirow{2}{*}{\shortstack{Inf. \\Size}} & \multicolumn{12}{c}{WER (\%)} & \multirow{2}{*}{\shortstack{Avg. \\WER (\%)}} \\ \cline{5-16}
     & & & & \multirow{1.2}{*}{en-US} & \multirow{1.2}{*}{fr-FR} & \multirow{1.2}{*}{es-US} & \multirow{1.2}{*}{es-ES} & \multirow{1.2}{*}{ja-JP} & \multirow{1.2}{*}{zh-TW} & \multirow{1.2}{*}{de-DE} & \multirow{1.2}{*}{it-IT} & \multirow{1.2}{*}{ar-EG} & \multirow{1.2}{*}{pt-BR} & \multirow{1.2}{*}{ru-RU} & \multirow{1.2}{*}{hi-IN} & \\[0.3ex] \hline\hline
    B1 & Multi. Cas. & 180M & 180M & 9.4 & 10.8 & 6.8 & 6.6 & 15.5 & 9.4 & 14.5 & 10.4 & 12.3 & 7.6 & 13.0 & 19.6 & 11.33 \\ \hline
    E2 & MoE-End & 400M & 211M & 8.3 & 9.3 & 6.0 & 4.9 & 12.9 & 9.1 & 12.6 & 8.8 & 10.8 & 6.9 & 10.9 & 19.2 & 9.98 \\ \hline
    B2 & Larger B1 & 400M & 400M & 8.1 & 9.1 & 6.2 & 5.4 & 14.5 & 9.3 & 12.4 & 8.1 & 10.7 & 6.4 & 10.6 & \textbf{19} & 9.98 \\ \hline
    B3 & B1+Adapter & 280M & 187M & \textbf{7.3} & 9.4 & 6.5 & 5.5 & 14.4 & \textbf{8.5} & 13.2 & 8.7 & \textbf{10.0} & 6.8 & 11.2 & 19.1 & 10.05 \\ \hline
    E8 & End-3-8 & 336M & 187M & 8.5 & 9.3 & 6.2 & 5.5 & 14.1 & 8.9 & 13.4 & 9.1 & 11.1 & 6.8 & 11.2 & 19.1 & 10.27 \\ \hline
    E9 & End-3-16 & 532M & 187M & 8.1 & 9.6 & 6.0 & 5.3 & 13.5 & 8.9 & 12.8 & 8.9 & 11.2 & 6.7 & 11.0 & 19.4 & 10.12 \\ \hline
    E10 & End-3-24 & 729M & 187M & 7.9 & 9.8 & 6.1 & 5.3 & 13.4 & 9.2 & 12.7 & 8.9 & 11.1 & 6.6 & 10.7 & 19.4 & 10.09 \\ \hline
    E11 & E2+SF & +128M LM & +128M LM & 8.1 & \textbf{8.0} & \textbf{5.7} & \textbf{4.8} & \textbf{12.4} & 10.4 & \textbf{11.6} & \textbf{7.7} & 10.6 & \textbf{6.1} & \textbf{10.3} & 20.4 & \textbf{9.68} \\ \hline
\end{tabular}
}
\caption{Comparison of multilingual baseline models, adapter models, and MoE models. The numbers in bold represent best WER for any language.}
\vspace{-0.4in}
\label{tab:wer_compare}
\end{table*}

\section{Results and Comparisons}
\label{sec:exp}

\subsection{Ablation Studies}
\label{sec:ablation}

\subsubsection{Place of MoE Layers}

In  Table  \ref{tab:wer_nc},  we add the MoE layers to different places of the Conformer layer and each MoE layer has 8 experts. We show in Table \ref{tab:wer_nc} that using MoE to replace the start FFN (i.e. MoE-Start) in a Conformer layer improves the baseline (B1) significantly by around 10.9\% relative on average.  We note the improvement is uniform and significant for all languages. To ablate on the location of where MoE layers are added, we also add MoE layers to the end FFN of the Conformer layer (\texttt{E2}, MoE-End), or both start and end FFNs (\texttt{E3}, MoE-Both). We see in Table \ref{tab:wer_nc} that using MoE at the end FFN is slightly better than at the start. Using MoE for both start and end FFN works best but it also increases the inference model size because we have more MoE layers at inference. We are aware that the improvement of MoE models in Table \ref{tab:wer_nc} may be due to the increased inference model size. In Sect. \ref{sec:compare}, we will compare to a larger baseline which has a similar size as the MoE model at the inference time.

We have also tried adding MoE layers to the causal encoder, and the average WER is significantly worse. In following sections, we use MoE-End given the slightly better performance and efficiency.

\subsubsection{Number of Experts}

To reduce the total model size, we tried varying the number of experts in the MoE layer in Table \ref{tab:wer_expert}. When we reduce the number of experts, the model total size decreases while the inference size stays the same because we always use the top 2 experts during inference. We see that the performance drops significantly when we decrease the expert number from 8 to 4, and relatively slowly from 4 to 2. This shows the model has been able to utilize the capacity from all experts. We also note that \texttt{E5} degenerates to a dense model and the performance difference between \texttt{B1} and \texttt{E5} is due to model size. We have also tried further increasing the number of experts and obtained better performance in Sect. \ref{sec:compare}.

\subsubsection{Reduce Number of MoE Layers}

Since reducing the number of experts reduces total model size but not inference model size, we have also tried reducing the number of MoE layers to make inference more efficient. In Table \ref{tab:wer_layer}, we reduce the number of MoE layers by only adding it to every other Conformer layer (MoE-End-Odd), or only the first Conformer layer (MoE-Conf1). We see that by reducing the number of MoE layers, the model performance drops significantly. However, we note that even adding one MoE layer at the first Conformer layer is helpful. Although it increases the baseline size by only 3M parameters, it reduces the average WER from 11.33\% to 10.88\%.

\subsection{Comparisons}
\label{sec:compare}

In Table \ref{tab:wer_compare}, we first compare the multilingual cascaded encoder baseline (\texttt{B1}) to the multilingual MoE-End model (\texttt{E2}), and \texttt{E2} reduces the average WER by 11.9\% relative compared by \texttt{B1}. We then increase the cascaded encoder of \texttt{B1} to a 896-D 17-layer Conformer to have a total size of 400M (i.e. \texttt{B2}), which is the same size as the multilingual MoE model (\texttt{E2}). We show that \texttt{B2} and \texttt{E2} have the same average WERs: 9.98\%, but the MoE model is around 47\% more efficient (211M vs 400M) in terms of model parameters activated for inference. We note that in practice, one needs to implement inference in a way that only activated experts are selected for forward propagation in order to achieve this efficiency.

We also compare to an adapter model (\texttt{B3}) which adds residual adapters \cite{houlsby2019parameter} to the baseline. The adapters we use follow a similar structure in \cite{kannan2019large, li2023efficient}, i.e., we insert a 512-D residual adapter after every Conformer layer in the cascaded encoder. In both training and decoding, ground truth language information is used to select the corresponding adapter. To compare to the adapter model fairly in inference (same activated parameters), we first reduce the FFN multiplier in the MoE layer from 4 to 3 and then increase the number of experts from 8, 16, to 24. The results in Table \ref{tab:wer_compare} show that when the number of experts increases from 8 to 16, the MoE model improves because of increased model capacity. When we increase the experts to 16 or 24, the MoE models (\texttt{E9}, \texttt{E10}) perform similarly to the adapter model on average (10.12\% or 10.09\% vs 10.05\%). However, we note that we do not explicitly use any language information in inference for MoE models while the adapter model uses ground truth language information. We also note that the improvement from 16-expert to 24-expert model (\texttt{E9} vs \texttt{E10}) is slight. This is probably because we only have 12 languages and the total number of experts may have exceed the needed capacity.

\subsection{Further Improvement by Shallow Fusion}

We further train a 128M multilingual LM by pooling the text portion of the supervised training data and text-only data from the 12 languages. Our text-only data covers 12 languages, and the sentences for each language ranges from 3.9B to 451B. 
The sentences are  sampled  from  anonymized  search  traffic  across multiple domains such as Web, Maps, News, Play, and YouTube.
We use a 12-layer Conformer LM for shallow fusion (SF). The Conformer LM has a model dimension of 768 and a feedforward layer dimension of 2048. A left context of 31 tokens is used to attend to the previous tokens. We use 6 heads for self-attention. The total model size is around 140M. We use the same wordpiece model as the MoE model.

As shown in Table \ref{tab:wer_compare}, shallow fusion further improves WER for almost all languages. The average WER reduction is around 3\% relative, with the largest improvement of around 14.0\% for fr-FR. We got regression for a couple of languages: zh-TW and hi-IN. The regression on zh-TW is probably because some of our text-only data contain Cantonese transcripts which is different from zh-TW. As for hi-IN, it has the largest amount of text-only data (around 451B sentences) and we may need to research filtering technique to better match the 
Search
domain.

\section{Conclusion}

We propose a streaming multilingual Conformer model with mixture-of-expert (MoE) layers.
By adding the MoE layers to the end FFN network of the Conformer, we improve the average WER of 12 languages by 11.9\% relative compared to the baseline. The proposed model is also efficient: Activating only 53\% of parameters during inference compared to a large baseline with similar quality. We also achieve similar quality compared to a ground truth language information based adapter model with increased experts but activating the same number of parameters in inference and not using language information. Further improvements have been obtained by multilingual shallow fusion.

\bibliographystyle{IEEEtran}
\bibliography{refs}

\end{document}